# GWO-FI: A novel machine learning framework by combining Gray Wolf Optimizer and Frequent Itemsets to diagnose and investigate effective factors on In-Hospital Mortality and Length of Stay among Kermanshahian Cardiovascular Disease patients

Ali Yavari,  Parisa Janjani, Sayeh Motavaseli, Seyran Weysi, Soraya Siabani, Mohammad Rouzbahani[*]

*Cardiovascular Research Center, Health Research Institute,* Imam Ali Hospital, *Kermanshah University of Medical Sciences, Kermanshah, Iran*
Email: Dr_ruzbahani78@yahoo.com

**Abstract**
Investigation and analysis of patient outcomes, including in-hospital mortality and length of stay, are crucial for assisting clinicians in determining a patient's result at the outset of their hospitalization and for assisting hospitals in allocating their resources. This paper proposes an approach based on combining the well-known gray wolf algorithm with frequent items extracted by association rule mining algorithms. First, original features are combined with the discriminative extracted frequent items. The best subset of these features is then chosen, and the parameters of the used classification algorithms are also adjusted, using the gray wolf algorithm. This framework was evaluated using a real dataset made up of 2816 patients from the Imam Ali Kermanshah Hospital in Iran. The study's findings indicate that low Ejection Fraction, old age, high CPK values, and high Creatinine levels are the main contributors to patients' mortality. Several significant and interesting rules related to mortality in hospitals and length of stay have also been extracted and presented.
Additionally, the accuracy, sensitivity, specificity, and auroc of the proposed framework for the diagnosis of mortality in the hospital using the SVM classifier were 0.9961, 0.9477, 0.9992, and 0.9734, respectively. According to the framework's findings, adding frequent items as features considerably improves classification accuracy.

## 1- Introduction

Cardiovascular diseases were the world's deadliest diseases prior to the COVID-19 pandemic's breakout [1]. However, these illnesses can become much more deadly in the years after the epidemic. Because quarantines have a considerable long-term impact on the incidence of heart disorders [2]. As a result, it is crucial to investigate and diagnose clinical data related to cardiac illnesses. For instance, the investigation of factors affecting a patient's length of stay facilitates hospitals in allocating hospital resources like the intensive care unit (ICU) or medical equipment, and the diagnosis of in-hospital mortality for a newly arrived patient using their medical data can help doctors make medical decisions about the outcome of these patients.
Han et al. demonstrated in [3, 4] that the use of frequent items increases the performance of diagnostic models. As an illustration, imagine that you are the Chief Executive Officer (CEO) of a bank and that you use the features of "credit worthy" and "high income" to distinguish loans to your clients. Of course, there can be a customer who violates one of the two rules; this implies that either he is a creditor and has a low income or he is not a creditor and has a large income, thus giving loans to those people is the wrong move. However, if these two features are combined to form a new feature called "good credit and high income," in our example, this feature has a much greater differentiation and will aid you in making the right choice. Similarly, by having hypertension and typical chest pain, figure 1 can be considered as a comparable point regarding medical conditions. However, the combination of these two features can create a powerful and unique feature called "hypertension and chest pain," which helps doctors in the diagnosis process in addition to the construction of a powerful diagnostic model. Separately, none of these features may be a strong cause of heart disease.

|   | HTN | ChestPain | HTN ∧ ChestPain |
|---|-----|-----------|-----------------|
| 1 | yes | yes | yes |
| 2 | yes | yes | yes |
| 3 | no  | yes | no  |
| 4 | yes | no  | no  |
| 5 | no  | no  | no  |
|   | ... | ... | ... |

**Figure 1: A simple example of combining itemsets**

In this paper, a potent framework for diagnosis and investigating in-hospital mortality besides discovering length of stay is proposed by combining the Gray Wolf Optimizer (GWO) algorithm and frequent itemsets extracted by Association Rule Mining (ARM) techniques. The gray wolf algorithm is used to optimize the number of optimal features needed for classification in addition to tuning a shallow and simple classifier, such as SVM, because the number of frequent itemsets extracted by ARM algorithms may be huge. The findings demonstrate that the use of shallow classifiers led to the creation of a very potent diagnostic system that uses a powerful and appealing feature extraction technique. This system accurately predicts the mortality conditions of the hospital. In this case, deep learning algorithms are not necessary because of how slowly it works.

We can notably mention examples [5-8, 11–15, 17–19] among the earlier efforts in the area of artificial intelligence (or machine learning) in medicine found in cases [5-23]. These methods try to model and forecast patient mortality and length of stay. In [18], language models like BioBERT were used to predict in-hospital mortality and length of stay using unstructured data, including clinical medical notes. Big medical data and artificial intelligence were combined by Soffer et al. [11] to build a mortality prediction model for hospitalized patients. Additionally, ML techniques like Random Forest in [6] have been utilized to find the two previously indicated targets. Convolutional networks were applied to the time series data of the MIMIC III dataset by Chen and his colleagues [7].

The following is a summary of this article's key innovations:

1. Proposing a very powerful and discriminative technique for feature extraction from medical data by frequent itemsets.
2. Combining the feature extraction method and the potent GWO algorithm to enhance the diagnostic system's performance.
3. Providing great performance to predict in-hospital mortality via the proposed framework which uses shallow classifiers with far fewer calculations than deep neural network techniques.
4. Investigation of effective factors on in-hospital mortality and length of stay, including used medications, medical history, etc., by interpretation of extracted association rules.
5. Evaluating the proposed framework by a real-world heart dataset.

The remainder of this article is organized as follows. Our real-world heart dataset is presented in Section 2. A description of the GWO-FI framework is then provided in Section 3. The results of implementing the proposed framework on the dataset are described in Section 4. After that, the most significant single effective factors on In-Hospital Mortality besides analyzing the effectiveness of medications on the patients are discussed in Section 5. In the end, a short conclusion is given in Section 6.

## 2- Iranian Imam Ali (AS) Kermanshah Hospital cardiac dataset

All adult STEMI patients older than 18 who visited Imam Ali Kermanshah Hospital in Iran between July 1, 2016, and December 21, 2019, were included in this study. Those with a confirmed STEMI diagnosis were hospitalized based on the cardiologist's recommendation and current guidelines [24]. Patients with STEMI who were admitted to Imam Ali (AS) Hospital for a different reason and then had a heart attack were not included, nor were patients with heart attacks who had been hospitalized in other medical facilities more than 24 hours earlier. A formal informed consent form was signed by each participant. The study protocol was accepted by the Kermanshah University of Medical Sciences ethics committee (ethical registration code: (KUMS.REC.1395.252). A checklist containing demographic data, medical records, cardiovascular risk factors, and the results of clinical examinations, diagnostic tests, treatment, and the course of the disease was completed by trained nurses at the time of arrival, throughout the hospitalization, and at the time of discharge in the systematic registration program for STEMI patients at Imam Ali Hospital (AS). The physician, who is a research associate, is in charge of ensuring the accuracy of the results that the nurses record. The complete list of these features is shown in Table 1. (laboratory findings were measured in the first 24 hours of admission). Fortunately, the data only had a low number of missing values, which the renowned MICE (Multivariate Imputation by Chained Equations) technique calculated. Additionally, in order to enrich the dataset, some continuous features' discrete versions were added in addition to their numerical versions in accord with the reference's normal values [25], as shown in Table 2.

**Table 1: List of features used in this research**

| Feature name | Range | Mean and standard deviation (std) / Frequency |
|---|---|---|
| Glucose plasma (mg/dl) | 60-597 | Mean: 156.64, std:76.71 |
| Age (years) | 19-94 | Mean: 60.74, std: 12.34 |
| Age2 (categorized of age) | young: if 18<age<30, middle-aged: if age<45 and male or age<55 and female, old: if age>45 and male or age>55 and female | Young: 20, middle-aged: 346, old: 2450 |
| Age3 (categorized of age) | 18-30, 31-40, 41-50, 51-60, >60 | 18-30: 20, 31-40: 105, 41-50: 470, 51-60: 807 , >60:1414 |
| BMI (body mass index) (Kg/m$^2$) | 13.88-65.54 | Mean: 26.23, std: 4.05 |
| CatBMI (categorized of BMI) | Underweight: if BMI<18.5, normal weight: if 18.5<=BMI<24.9, Overweight: if 25<=BMI<29.9, Obesity class1: if 30<=BMI<34.9, Obesity class 2: if 35<=BMI<39.9 Obesity class3: >40 | Underweight: 48, normal weight: 1078, Overweight: 1240, Obesity class1: 378, Obesity class2:61, Obesity class3: 11 |
| Obesity | Yes if MBI > 25, no otherwise | Yes: 1690, no: 1126 |
| Early Creatinine (mg/dl) | 0.3-11.8 | Mean: 1.162, std: 0.424 |
| Highest Creatinine (mg/dl) | 0.6-23 | Mean: 1.3, std: 0.77 |
| Systolic Blood Pressure (mmHg) | 40-250 | Mean: 134.11, std: 29.88 |
| GFR (Glomerular Filtration Rate) (mL/min) [26] | 3.35-129.65 | Mean: 68.39, std: 18.19 |
| CatGFR (categorized) | Normal: if GFR>60, kidney disease: if 15<=GFR<60, kidney failure: if GFR<15 | Normal: 1928, kidney disease: 880, kidney failure:8 |
| Last EF (ejection fraction) (%) | 10-60 | Mean: 37.69, std: 9.61 |
| Early HB (hemoglobin) (g/dl) | 7.4-23.9 | Mean: 14.73, std: 1.80 |
| Least HB (hemoglobin) (g/dl) | 6-20.9 | Mean: 13.80, std: 2.239 |
| PLT (platelet) (1000/ml) | 38-928 | Mean: 243.25, std: 73.03 |
| WBC (white blood cell) (cells/ml) | 1.1-35.5 | Mean: 11.14 , std: 3.582 |
| LDL (low density lipoprotein) (mg/dl) | 21-295 | Mean: 103.983, std: 30.37 |
| HDL (high density lipoprotein) (mg/dl) | 14-98 | Mean: 41.34, std: 8.913 |
| TG (triglyceride) (mg/dl) | 26.40-1850 | Mean: 142.29, std: 89.72 |
| Total Cholestrol (mg/dL) | 80-510 | Mean: 175.06, std: 40.76 |
| First CPK (creatine phosphokinase) (units/L) | 4.2-18000 | Mean: 1567.48, std: 1489.26 |
| Highest CPK (creatine phosphokinase) (units/L) | 50-18000 | Mean: 1801.84, std: 1589.12 |
| First CKMB (IU/L) | 4.39-530 | Mean: 29.153, std: 40.641 |
| Highest CKMB (IU/L) | 11-2449 | Mean: 126.01, std: 118.919 |
| Heart Rate in First EKG (ppm) | 18-214 | Mean: 78.18, std: 20.40 |
| Time between symptom onset and admission – Other hospitals (h) | 0.083-729.3 | Mean: 3.85, std: 14.94 |
| Time between symptom other(categorized) | <4, 4-8, >8 | <4: 2111, 4-8: 430, >8: 275 |
| Time between symptom onset and admission – Only Imam Ali (h) | 0.083-754.66 | Mean: 5.46, std: 25.25 |
| Time between symptom only Imam Ali (categorized) | <4, 4-8, >8 | <4: 1829 , 4-8:579, >8: 408 |
| Gender | Female, male | Male: 2181, female: 635 |
| Education | Illiterate, elementary, 6-12, diploma, bachelor, associate degree, master, doctorate | Illiterate: 854, elementary: 675, 6-12: 490, diploma: 492, associate degree: 102, bachelor: 139 , master: 53, doctorate: 11 |
| Place | Kermanshah city, other cities, villages | Kermanshah: 2177, other cities: 323, villages:316 |
| First Troponin (binary) | Yes, no | Yes: 391, no: 2425 |
| Higest Troponin (binary) | Yes, no | Yes: 2740, no: 76 |

| Feature name | Range | Mean and standard deviation (std) / Frequency |
|---|---|---|
| Anterior MI/LBBB | Yes, no | Yes: 617, no: 2199 |
| History of Coronary | Yes, no | Yes: 536, no: 2280 |
| History of Intervention | Yes, no | Yes: 256, no: 2560 |
| History of Angina | Yes, no | Yes: 476, no: 2340 |
| Current Smoker | Yes, no | Yes: 1373, no: 1443 |
| Diabetes | Yes, no | Yes: 595, no: 2221 |
| History of HLP (hyperlipidemia) | Yes, no | Yes: 665, no: 2151 |
| THTN (hypertension) | Yes, no | Yes: 1210, no: 1606 |
| Life limiting disease | Yes, no | Yes: 648, no: 2168 |
| Type Of FMC | 1, 2, 3, 4 | 1: 290, 2: 322, 3: 4, 4: 2200 |
| Admission Mode | Theirselves, ambulance | Theirselves: 2465, amnulance: 351 |
| Direct admission to Imam-Ali hospital | Yes, no | Direct: 2079, Indirect: 737 |
| Out-of-hospital cardiac arrests | Yes, no | Yes: 56, no: 2760 |
| Ventilation options | Yes, no | Yes: 83, no: 2733 |
| WorstKLLIPclass | 1,2,3,4 | 1: 1997, 2: 435, 3: 69, 4: 315 |
| Bleading | 1, 2, 3, 4, 5, 6, 7 | 0: 2462, 1-7: 354 |
| PCI in hospital (Second PCI) | Yes, no | Yes: 69, no: 2747 |
| CABG in Hospital | Yes, no | Yes: 241, no: 2575 |
| Reperfusion therapy | PPCI: 1, Pharmaco-invasive: 2, Thrombolytic drug: 3, Without reperfusion therapy: 4 | 1: 1639, 2: 320, 3: 421, 4: 436 |
| Transfusion | Yes, no | Yes: 228, no: 2588 |
| Heart Failure (HF) in hospital | Yes, no | Yes: 692, no: 2124 |
| Atrial fibrillation (AF) in First EKG | Yes, no | Yes: 74, no: 2742 |
| Atrial fibrillation (AF) in Hospital | Yes, no | Yes: 140, no: 2676 |
| Aspirin | Yes, no | Yes: 661, no: 2155 |
| AspirinHos (Aspirin in hospital) | Yes, no | Yes: 2807, no: 9 |
| Clopidogrel | Yes, no | Yes: 83, no: 2733 |
| ClopidogrelHos (Clopidogrel in hospital) | Yes, no | Yes: 2740, no: 76 |
| Eptifibatide | Yes, no | Yes: 2, no: 2814 |
| EptifibatideHos (Eptifibatide in hospital) | Yes, no | Yes: 1126, no: 1690 |
| heparinunfrac | Yes, no | Yes: 2, no: 2814 |
| heparinunfracHos (heparinunfrac in hospital) | Yes, no | Yes: 2069, no: 747 |
| heparinLW | Yes, no | Yes: 6, no: 2810 |
| heparinLWHos (heparinLW in hospital) | Yes, no | Yes: 1221, no: 1595 |
| BetaBlockers | Yes, no | Yes: 463, no: 2353 |
| BetaBlockersHos (BetaBlockers in hospital) | Yes, no | Yes: 2528, no: 288 |
| ACEinhibitors | Yes, no | Yes: 254, no: 2562 |
| ACEinhibitorsHos (ACEinhibitors in hospital) | Yes, no | Yes: 2570, no: 246 |
| ARBs | Yes, no | Yes: 587, no: 2229 |
| ARBsHosp (ARBs in hospital) | Yes, no | Yes: 227, no: 2589 |
| MRAs | Yes, no | Yes: 44, no: 2772 |
| MRAsHosp (MRAs in hospital) | Yes, no | Yes: 595, no: 2221 |
| Digoxin | Yes, no | Yes: 16, no: 2800 |
| DigoxinHosp (Digoxin in hospital) | Yes, no | Yes: 78, no: 2738 |
| Diuretics | Yes, no | Yes: 144, no: 2672 |
| DiureticsHosp (Diuretics in hospital) | Yes, no | Yes: 755, no: 2061 |
| Statins | Yes, no | Yes: 357, no: 2459 |
| StatinsHosp (Statins in hospital) | Yes, no | Yes: 2792, no: 21 |
| PPIs | Yes, no | Yes: 48, no: 2768 |
| PPIsHos (PPIs in hospital) | Yes, no | Yes: 2578, no: 238 |
| Length of stay (day) | 0-96 | Mean: 6.34, Mean: 6.64 |
| Cat Length of stay2 (categorized of Length of stay) | <3, 3-7, 7-14, >14 | <3: 318, 3-7: 1723, 7-14: 497, >14: 278 |
| In-Hospital Mortality | Yes, no | Yes: 172, no: 2644 |

**Table 2: List of created categorical features used in this research**

| Feature name | Gender | Low | Normal | High |
|---|---|---|---|---|
| Glucose plasma2 | - | <60 | 60-200 | >200 |

| Feature name | Gender | Low | Normal | High |
|---|---|---|---|---|
| EarlyCr2 and Highest Cr | Male | <0.75 | 0.75-1.2 | >1.2 |
| | Female | 0.65 | 0.65-1 | >1 |
| WBC2 | - | <4000 | 4000-10000 | >10000 |
| Systolic Blood Pressure2 | - | <90 | 90-140 | >140 |
| Last EF2 (EF2) | - | <=50 | >50 | |
| LDL2 | - | - | <=130 | >130 |
| HDL | - | - | <=40 | >40 |
| TG2 | - | - | <=200 | >200 |
| Total Cholestrol2 | - | - | <=200 | >200 |
| First CPK2 and highest CPK2 | - | <200 | 20-200 | >200 |
| Early HB2 and Least HB2 | Male | <13.5 | 13.5-17.5 | >17.5 |
| | Female | <12 | 12-16 | >16 |
| First CKMB2 and Highest CKMB2 | - | <5 | 5-25 | >25 |
| PLT2 | - | <150 | 150-399 | >399 |
| Heart Rate in First EKG2 | - | <60 | 60-100 | >100 |

## 3- GWO-FI: A novel framework by combining Gray Wolf Optimizer and Frequent Itemsets to diagnose In-Hospital Mortality

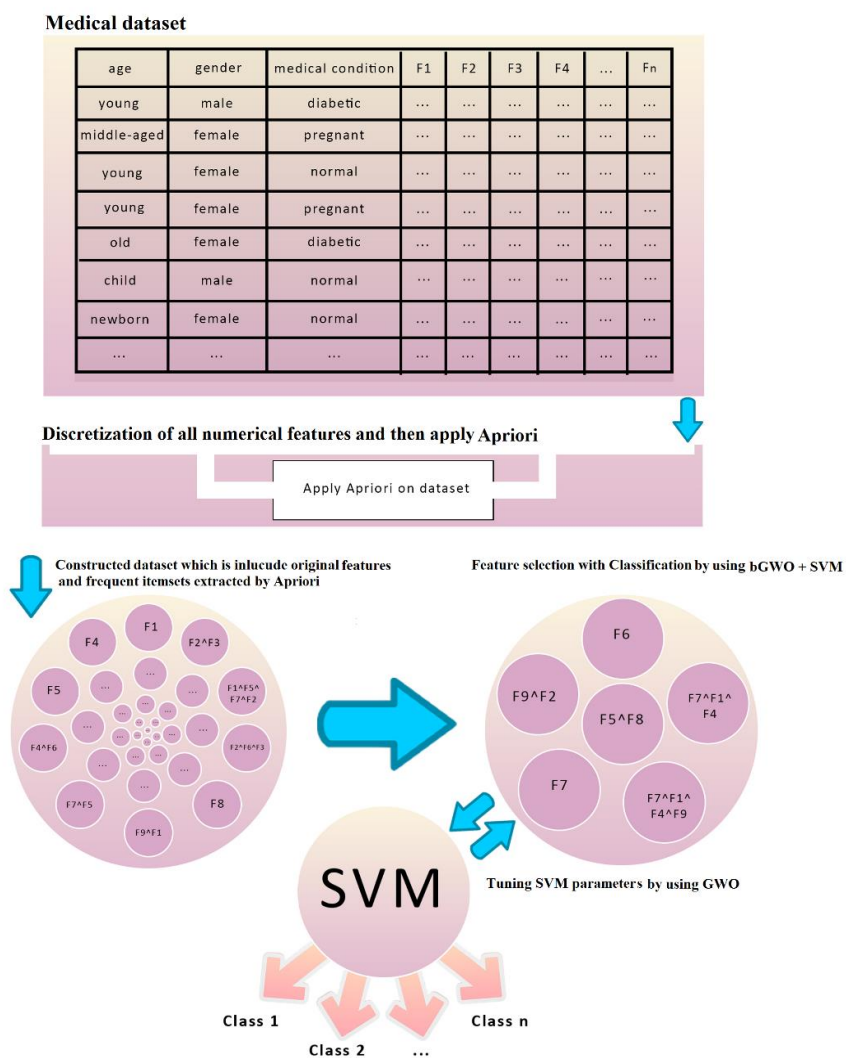

**Figure 2:** Scheme of the proposed method: Combining Gray Wolf Optimizer and Frequent Itemsets to diagnose In-Hospital Mortality

The Gray Wolf Optimizer (GWO), a swarm intelligence-related optimization algorithm developed by Mirjalili et al. in 2014 [27], mimics the hunting behavior of gray wolves (Canis lupus) in the wild. During this hunt, wolves follow the leader and hunter, the Alpha Wolf. There are three key steps to this hunting: looking for prey, surrounding the prey, and attacking the prey. This hunting was done in a flock with four different sorts of grey wolves: alpha, beta, delta, and omega. The binary Gray Wolf Optimizer (bGWO), a potent discrete variant of this approach that was published in 2016, is employed in this study to choose the best feature subset [28]. The proposed framework, GWO-FI, for predicting in-hospital mortality is shown schematically in Figure 2. There are numerous ARM techniques for extracting frequent itemsets, with Apriori being one of the most well-known [29]. In this article, we extract frequent itemsets of length k and add them to the original dataset using the first step of the apriori technique. Then, using bGWO and a shallow classifier that we trained, we try to choose the best subset of features. It should be highlighted that GWO is also used to optimize the parameters of this shallow method (continuous version). The reason for using the optimization algorithm in this problem is that a very large number of frequent itemsets are produced whenever the value of the control parameter of phase 1 of the apriori algorithm, i.e., support, increases, and using all of them for the classification task may not have desirable performance. However, the availability of a shallow classifier for scoring each subset produced by GWO makes it possible to choose the optimal subset. Additionally, utilizing the second apriori stage, strong rules are developed in Section 4-2 in order to examine the factors affecting the targets.

**3-1 Evaluation measures of the proposed framework**

Accuracy, sensitivity, and specificity measurements are used in this research to evaluate the classification algorithm in accordance with formulas 1 to 3. The area under the receiver operating characteristic (auroc) measure is utilized as a single evaluation metric due to the imbalanced nature of the dataset for the problem of in-hospital mortality, i.e., the fact that there are roughly 15 times as many samples of the negative classes as there are positive classes. It is capable of accurately measuring the diagnostic system's performance.

$$\text{Accuracy} = \frac{Tp + Tn}{Fp + Fn + Tp + Tn} \quad (1)$$

$$\text{Sensitivity} = \frac{Tp}{Tp + Fn} \quad (2)$$

$$\text{Speceficity} = \frac{Tn}{Tn + Fp} \quad (3)$$

On the other hand, ARM algorithms evaluate frequent items and rules using two key support and confidence measures (equations 4 and 5). However, as was previously stated, since the dataset for the issue of in-hospital mortality is imbalanced, we additionally utilize max confidence, Kuczynski, cosine, and imbalance ratio measures (equations 6–9; null-invariant measures) to find more interesting rules in addition to the two major measures [3].

$$sup(A.B) = \frac{count(A)}{total\ number\ of\ patients} \quad (4)$$

$$conf(A \rightarrow B) = \frac{sup(A \cup B)}{sup(A)} \quad (5)$$

$$max\ confidence(A.B) = \max\{conf(A \rightarrow B).conf(B \rightarrow A)\} \quad (6)$$

$$kulczynski(A.B) = \frac{conf(A \rightarrow B) + conf(B \rightarrow A)}{2} \quad (7)$$

$$cosine(A.B) = \frac{sup(A \cup B)}{\sqrt{sup(A) + sup(B)}} \quad (8)$$

$$IR(A.B) = \frac{|\sup(A) - \sup(B)|}{\sup(A) + \sup(B) - \sup(A \cup B)} \tag{9}$$

## 4- Evaluation and results

This section explains the results of implementing the proposed method on our real-world heart dataset, including the factors impacting in-hospital mortality and length of stay as well as the in-hospital mortality diagnosis system.

### 4-1 In-Hospital Mortality diagnosis results

The diagnostic model created for predicting in-hospital mortality is shown in Table 3. SVM, Naive Bayes, and Decision Tree, three shallow classifiers, were employed in this experiment. The SVM classifier produced the best result, according to the findings, with a remarkable accuracy of 99.61% and rates of 0.9477 for sensitivity, 0.9992 for specificity, and 0.9734 for auroc (only 9 samples were misclassified). It is clear that the results of classifiers trained by the Gray Wolf algorithm without using the proposed feature extraction method are ineffective and unable to accurately predict the samples; in particular, these classifiers (except for Naïve Bayes) are unable to identify positive cases (fatality) or mortality within the hospital (low specificity and therefore low auroc). This is because the proposed framework (which combines the proposed feature extraction method and the gray wolf feature selection algorithm) extracts and then selects very discriminative features. The effectiveness of the Naïve Bayes and Decision Tree classifiers has improved when the proposed framework and SVM are combined.

Table 3: The results of built classifiers to predict In-Hospital Mortality with/without using the GWO-FI framework

| Method | #selected features | Accuracy (%) | Sensitivity | Specificity | Auroc |
|---|---|---|---|---|---|
| **Proposed method: GWO + Apriori + SVM** | **97** | **99.61** | **0.9477** | **0.9992** | **0.9734** |
| **Proposed method: GWO + Apriori + Decision Tree** | 94 | 98.83 | 0.9419 | 0.9913 | 0.9665 |
| **Proposed method: GWO + Apriori + Naïve Bayes** | 131 | 96.09 | 0.9709 | 0.9603 | 0.9656 |
| Original dataset + GWO + SVM | 49 | 98.22 | 0.8663 | 0.9898 | 0.9280 |
| Original dataset + GWO + Decision Tree | 56 | 97.44 | 0.8779 | 0.9807 | 0.9293 |
| Original dataset + GWO + Naïve Bayes | 39 | 92.86 | 0.9651 | 0.9262 | 0.9456 |

### 4-2 The most interesting effective factors on In-Hospital Mortality and Length of stay

Understanding and investigating the causes of an event can have a great impact on preventing it again. In this section, a variety of factors impacting the mortality in the hospital as well as the length of hospitalization are extracted with the aid of the significant and frequently used apriori algorithm.

The most interesting effective factors on in-hospital mortality are listed in Table 4 and are sorted according to the Max Confidence measure. A large number of the rules in this table are interpretable and explainable. For example, according to the latest rule, if the patient is old and does not have a good heart condition (WorstKLLIPclass=four), then there is a high probability of death. This rule is true for 152 patients who died (out of 172 patients who died). Or according to another rule, 45 patients with improper heart conditions who were overweight and did not receive any treatment died. This death may have happened in the first hours when the patient did not have the chance to receive any kind of treatment. Also, 46 people with poor heart conditions who were connected to ventilators and had heart attacks in the hospital died with 90% confidence.

Although the length of stay is a target in our problem, in order to investigate this important point that the length of stay can have an effect on in-hospital mortality, it is also considered as one of the features in this analysis. For example, according to the first rule of Table 4, 74 people who had a high early blood creatinine, had inappropriate heart conditions, and stayed in the hospital for less than 3 days died with 98.66% confidence. Also, according to the third, fourth, and fifth rules of this table, those who stayed in the hospital for less than three days and did not have appropriate heart conditions, and also those who were over 60 years old, had kidney failure, and high blood pressure, respectively, with high confidence, have died. In the fifth part, a complete discussion and consideration of drugs and their effects on patients have been done.

**Table 4: The most important factors affecting In-Hospital Mortality**

| Rules (X → Y) | Support | count | confidence | Max Confidence | Kuczynski | cosine | imbalance |
|---|---|---|---|---|---|---|---|
| {WorstKLLIPclass=four,ACEinhibitorsHos=no,LOS2=<3} => {InhospitalMortality=inhospital death} | 0.0156 | 44 | 1 | 1 | 0.6279 | 0.5057 | 0.7441 |
| {Higestcr2=high,WorstKLLIPclass=four,LOS2=<3} => {InhospitalMortality=inhospital death} | 0.0333 | 94 | 0.9894 | 0.9894 | 0.7679 | 0.7353 | 0.4450 |
| {EarlyCr2=high,WorstKLLIPclass=four,LOS2=<3} => {InhospitalMortality=inhospital death} | 0.0262 | 74 | 0.9866 | 0.9866 | 0.7084 | 0.6515 | 0.5606 |
| {WorstKLLIPclass=four,Education=illiterate,LOS2=<3} => {InhospitalMortality=inhospital death} | 0.0177 | 50 | 0.9803 | 0.9803 | 0.6355 | 0.5338 | 0.6994 |
| {Age3=61-above,WorstKLLIPclass=four,LOS2=<3} => {InhospitalMortality=inhospital death} | 0.0276 | 78 | 0.975 | 0.975 | 0.7142 | 0.6649 | 0.5287 |
| {GFR2=kidney_disease,WorstKLLIPclass=four,LOS2=<3} => {InhospitalMortality=inhospital death} | 0.0266 | 75 | 0.9740 | 0.9740 | 0.7050 | 0.6517 | 0.5459 |
| {THTN=yes,WorstKLLIPclass=four,LOS2=<3} => {InhospitalMortality=inhospital death} | 0.0202 | 57 | 0.9661 | 0.9661 | 0.6487 | 0.5658 | 0.6494 |
| {STDminonlyAli_SMh2=<4,WorstKLLIPclass=four,LOS2=<3} => {InhospitalMortality=inhospital death} | 0.0181 | 51 | 0.9622 | 0.9622 | 0.6293 | 0.5341 | 0.6839 |
| {AdmissMod=self,WorstKLLIPclass=four,LOS2=<3} => {InhospitalMortality=inhospital death} | 0.0308 | 87 | 0.9560 | 0.9560 | 0.7309 | 0.6953 | 0.4602 |
| {HDL2=low,WorstKLLIPclass=four,LOS2=<3} => {InhospitalMortality=inhospital death} | 0.015 | 43 | 0.9555 | 0.9555 | 0.6027 | 0.4887 | 0.7298 |
| {TypeOf_FMC=four,WorstKLLIPclass=four,LOS2=<3} => {InhospitalMortality=inhospital death} | 0.0301 | 85 | 0.9550 | 0.9550 | 0.7246 | 0.6870 | 0.4715 |
| {WorstKLLIPclass=four} => {InhospitalMortality=inhospital death} | 0.0578 | 163 | 0.5174 | 0.9476 | 0.7325 | 0.7002 | 0.4413 |
| {EF2=low,WorstKLLIPclass=four} => {InhospitalMortality=inhospital death} | 0.0571 | 161 | 0.5193 | 0.9360 | 0.7277 | 0.6972 | 0.4299 |
| {WorstKLLIPclass=four,Second_PCI=no} => {InhospitalMortality=inhospital death} | 0.0568 | 160 | 0.5211 | 0.9302 | 0.7257 | 0.6962 | 0.4231 |
| {Higest_CPK2=high,WorstKLLIPclass=four} => {InhospitalMortality=inhospital death} | 0.0561 | 158 | 0.5231 | 0.9186 | 0.7208 | 0.6932 | 0.4113 |
| {EF2=low,Higest_CPK2=high,WorstKLLIPclass=four} => {InhospitalMortality=inhospital death} | 0.0553 | 156 | 0.5234 | 0.9069 | 0.7152 | 0.6890 | 0.4012 |
| {VENTOption=yes,WorstKLLIPclass=four,HF_in_hospit=yes} => {InhospitalMortality=inhospital death} | 0.0163 | 46 | 0.9019 | 0.9019 | 0.5847 | 0.4911 | 0.6836 |
| {Higest_CPK2=high,WorstKLLIPclass=four,Second_PCI=no} => {InhospitalMortality=inhospital death} | 0.0550 | 155 | 0.5272 | 0.9011 | 0.7141 | 0.6892 | 0.3922 |
| {CatBMI=Overweight,WorstKLLIPclass=four,mainfourcat_sm=four} => {InhospitalMortality=inhospital death} | 0.0159 | 45 | 0.9 | 0.9 | 0.5808 | 0.4852 | 0.6892 |
| {Age2=old,WorstKLLIPclass=four} => {InhospitalMortality=inhospital death} | 0.0539 | 152 | 0.5241 | 0.8837 | 0.7039 | 0.6805 | 0.3806 |

The list of the most interesting and significant rules related to the length of hospitalization is presented in Table 5 below. The length of stay has been converted into a discrete variable of four values, as shown in Table 1. This table is sorted by the max confidence value for each group. First of all, it should be noted that 106 out of the 172 people who passed away were hospitalized for less than three days. This indicates that, regrettably, 61% of them died within the first three days (318 out of 2816 people); 45 of these 106 people received no treatment, which may have indicated their urgent condition and predicted death. Additionally, 69 of the deceased who arrived at the hospital without an ambulance, had kidney disease, and died in the hospital were hospitalized for less than three days.

But it should be noted that patients who spent between three and seven days in the hospital make up the largest group of study participants (1723 people); 50 of these patients had high levels of early haemoglobin and glucose plasma; 1709 of them did not require transfusions; 1674 did not bleed; and 1707 did not require a second PCI procedure. 24 overweight patients who arrived at the hospital on their own, had a second PCI, and spent 7–14 days in the hospital (497 patients) fall into this category. Additionally, 15 of those with a high first CPK, a history of blood lipids, and a second PCI in the hospital spent at least one week in the hospital. Finally, we can say that 44 of the patients who received the second kind of treatment (pharmaco-invasive) and in hospitals that have had CABG belong to the category of patients who were hospitalised for more than 14 days (278 people; the lowest category). Additionally, 75 patients who had grade 6 hemorrhage and hypertension were hospitalized for at least 14 days despite having no negative hospital events.

**Table 5: The most important factors affecting length of stay**

| Rules (X → Y) | Cat | Support | count | confidence | Max Confidence | Kulczynski | cosine | imbalance |
|---|---|---|---|---|---|---|---|---|
| {EF2=low,BetaBlockersHosp=no,InhospitalMortality=inhospital death} => {LOS2=<3} | <=3 | 0.01778 | 50 | 0.8620 | 0.8621 | 0.5097 | 0.3682 | 0.7975 |
| {GFR2=kidney_disease,WorstKLLIPclass=four,BetaBlockersHos=no} => {LOS2=<3} | <=3 | 0.0145 | 41 | 0.8541 | 0.8541 | 0.4915 | 0.3318 | 0.8307 |
| {AdmissMod=self,BetaBlockersHos=no,InhospitalMortality=inhospital death} => {LOS2=<3} | <=3 | 0.0152 | 43 | 0.8431 | 0.8431 | 0.4891 | 0.3376 | 0.819 |
| {Glocoseplasma2=high,WorstKLLIPclass=four,BetaBlockersHosp=no} => {LOS2=<3} | <=3 | 0.0166 | 47 | 0.7833 | 0.7833 | 0.4655 | 0.3402 | 0.7794 |
| {EF2=low,WorstKLLIPclass=four,BetaBlockers2=no} => {LOS2=<3} | <=3 | 0.0177 | 50 | 0.7575 | 0.7575 | 0.4574 | 0.3451 | 0.7544 |
| {Age2=old,WorstKLLIPclass=four,BetaBlockers2=no} => {LOS2=<3} | <=3 | 0.0170 | 48 | 0.75 | 0.75 | 0.4504 | 0.3364 | 0.7604 |
| {GFR2=kidney_disease,AdmissMod=self,InhospitalMortality=inhospital death} => {LOS2=<3} | <=3 | 0.0245 | 69 | 0.6764 | 0.6764 | 0.4467 | 0.3831 | 0.6153 |
| {InhospitalMortality=inhospital death} => {LOS2=<3} | <=3 | 0.0376 | 106 | 0.6162 | 0.6162 | 0.4748 | 0.4532 | 0.3802 |
| {mainfourcat_sm=four,InhospitalMortality=inhospital death} => {LOS2=<3} | <=3 | 0.0159 | 45 | 0.5625 | 0.5625 | 0.3520 | 0.2821 | 0.6742 |
| {Transfusion=no} => {LOS2=3to7} | 3to7 | 0.6068 | 1709 | 0.6603 | 0.9918 | 0.8261 | 0.8093 | 0.3324 |
| {Bleading=zero} => {LOS2=3to7} | 3to7 | 0.5945 | 1674 | 0.6799 | 0.9716 | 0.8257 | 0.8128 | 0.2943 |
| {Second_PCI=no} => {LOS2=3to7} | 3to7 | 0.6062 | 1707 | 0.6214 | 0.9907 | 0.8061 | 0.7846 | 0.3706 |
| {LeastHb2=high,Second_PCI=no,mainfourcat_sm=one} => {LOS2=3to7} | 3to7 | 0.0178 | 50 | 0.8772 | 0.8772 | 0.4531 | 0.1595 | 0.963 |
| {LeastHb2=high,CABGinHos=no,mainfourcat_sm=one} => {LOS2=3to7} | 3to7 | 0.0178 | 50 | 0.8621 | 0.8621 | 0.4455 | 0.1582 | 0.9619 |
| {EF2=low,LeastHb2=high,mainfourcat_sm=one} => {LOS2=3to7} | 3to7 | 0.0178 | 50 | 0.8621 | 0.8621 | 0.4455 | 0.1582 | 0.9619 |
| {EarlyHb2=high,Glocoseplasma2=high,mainfourcat_sm=one} => {LOS2=3to7} | 3to7 | 0.0178 | 50 | 0.8621 | 0.8621 | 0.4455 | 0.1582 | 0.9619 |
| {Age3=31-40,STDminAliornon_SMh2=<4,mainfourcat_sm=one} => {LOS2=3to7} | 3to7 | 0.0153 | 43 | 0.86 | 0.86 | 0.4425 | 0.1465 | 0.9671 |
| {Age3=41-50,AdmissMod=ambulance,mainfourcat_sm=one} => {LOS2=3to7} | 3to7 | 0.0153 | 43 | 0.86 | 0.86 | 0.4425 | 0.1465 | 0.9671 |
| {Age3=31-40,mainfourcat_sm=one,BetaBlockers2=yes} => {LOS2=3to7} | 3to7 | 0.0174 | 49 | 0.8448 | 0.8448 | 0.4366 | 0.155 | 0.9613 |
| {Higest_Tro_Bin=yes,AdmissMod=self,InhospitalMortality=alive} => {LOS2=3to7} | 3to7 | 0.5128 | 1444 | 0.6389 | 0.8381 | 0.7385 | 0.7318 | 0.2115 |
| {AdmissMod=self,ClopidogrelHosp=yes,InhospitalMortality=alive} => {LOS2=3to7} | 3to7 | 0.5107 | 1438 | 0.6368 | 0.8346 | 0.7357 | 0.729 | 0.2104 |
| {HigestCKMB2=high,pHLP=yes,Second_PCI=yes} => {LOS2=7to14} | 7to14 | 0.0053 | 15 | 0.7895 | 0.7895 | 0.4098 | 0.1544 | 0.9541 |
| {FirstCPK2=high,pHLP=yes,Second_PCI=yes} => {LOS2=7to14} | 7to14 | 0.0053 | 15 | 0.7895 | 0.7895 | 0.4098 | 0.1544 | 0.9541 |
| {Higest_CPK2=high,pHLP=yes,Second_PCI=yes} => {LOS2=7to14} | 7to14 | 0.0053 | 15 | 0.7895 | 0.7895 | 0.4098 | 0.1544 | 0.9541 |
| {Age3=51-60,Higest_CPK2=high,Second_PCI=yes} => {LOS2=7to14} | 7to14 | 0.0057 | 16 | 0.7619 | 0.7619 | 0.3970 | 0.1566 | 0.9482 |
| {pHLP=yes,Second_PCI=yes} => {LOS2=7to14} | 7to14 | 0.0053 | 15 | 0.75 | 0.75 | 0.3901 | 0.1505 | 0.9502 |
| {EF2=low,pHLP=yes,Second_PCI=yes} => {LOS2=7to14} | 7to14 | 0.0053 | 15 | 0.75 | 0.75 | 0.3901 | 0.1505 | 0.9502 |
| {Age3=51-60,HigestCKMB2=high,Second_PCI=yes} => {LOS2=7to14} | 7to14 | 0.0053 | 15 | 0.75 | 0.75 | 0.3901 | 0.1505 | 0.9502 |
| {Age3=51-60,FirstCPK2=high,Second_PCI=yes} => {LOS2=7to14} | 7to14 | 0.0053 | 15 | 0.75 | 0.75 | 0.3901 | 0.1505 | 0.9502 |
| {CatBMI=Overweight,AdmissMod=self,Second_PCI=yes} => {LOS2=7to14} | 7to14 | 0.0085 | 24 | 0.75 | 0.75 | 0.3991 | 0.1903 | 0.9208 |
| {CABGinHos=yes,mainfourcat_sm=two} => {LOS2=>14} | >14 | 0.0156 | 44 | 0.9362 | 0.9362 | 0.5472 | 0.3849 | 0.8221 |
| {CABGinHos=yes,mainfourcat_sm=two,BetaBlockersHosp=yes} => {LOS2=>14} | >14 | 0.0153 | 43 | 0.9348 | 0.9348 | 0.5447 | 0.3802 | 0.8256 |
| {THTN=yes,Bleading=six,InhospitalMortality=alive} => {LOS2=>14} | >14 | 0.0266 | 75 | 0.8241 | 0.8241 | 0.5469 | 0.4715 | 0.6360 |
| {Bleading=six,BetaBlockersHosp=yes,InhospitalMortality=alive} => {LOS2=>14} | >14 | 0.0558 | 157 | 0.8220 | 0.8220 | 0.6934 | 0.6813 | 0.2788 |
| {Bleading=six,mainfourcat_sm=four,Aspirin=no} => {LOS2=>14} | >14 | 0.0174 | 49 | 0.7903 | 0.7903 | 0.4833 | 0.3732 | 0.7423 |

## 5- Discussion
In this section, the single mortality-related factors are first extracted, and then the effectiveness of a few commonly used medications is assessed.

### 5-1 Extraction of single factors related to In-Hospital Mortality
We need to develop rules with a single item on the left side in order to extract the single factors that affect in-hospital mortality. Table 6 shows that of the 172 patients who passed away in hospitals, 170 had low Ejection Fractions (EF). For instance, 163 of them had improper cardiac function (WorstKLLIPclass=four). Additionally, 150 of them went to the hospital on their own. By calling the emergency, the family of these individuals may lessen the likelihood that they would pass away. 145 of them were transported to the hospital by ambulance with the emergency staff present, and 160 of them (as indicated in Table 1) were old. Some of them would have been saved from death if the doctor had been there.

Table 6: The most important single factors affecting In-Hospital Mortality

| Rules (X → Y) | Support | count | confidence | Max Confidence | Kuczynski | cosine | imbalance |
|---|---|---|---|---|---|---|---|
| {EF2=low} => {InhospitalMortality=inhospital death} | 0.0603 | 170 | 0.0622 | 0.9883 | 0.5252 | 0.2479 | 0.9363 |
| {Higest_CPK2=high} => {InhospitalMortality=inhospital death} | 0.0586 | 165 | 0.0630 | 0.9593 | 0.5112 | 0.2459 | 0.9318 |
| {WorstKLLIPclass=four} => {InhospitalMortality=inhospital death} | 0.0578 | 163 | 0.5174 | 0.9476 | 0.7325 | 0.7002 | 0.4413 |
| {Age2=old} => {InhospitalMortality=inhospital death} | 0.0568 | 160 | 0.0653 | 0.9302 | 0.4978 | 0.2465 | 0.9253 |
| {FirstCPK2=high} => {InhospitalMortality=inhospital death} | 0.0568 | 160 | 0.0633 | 0.9302 | 0.4968 | 0.2427 | 0.9275 |
| {Higestcr2=high} => {InhospitalMortality=inhospital death} | 0.0543 | 153 | 0.1332 | 0.8895 | 0.5113 | 0.3442 | 0.8365 |
| {pIntervention=no} => {InhospitalMortality=inhospital death} | 0.0543 | 153 | 0.0597 | 0.8895 | 0.4746 | 0.2305 | 0.9259 |
| {AdmissMod=self} => {InhospitalMortality=inhospital death} | 0.0532 | 150 | 0.0608 | 0.8720 | 0.4664 | 0.2303 | 0.9219 |
| {TypeOf_FMC=four} => {InhospitalMortality=inhospital death} | 0.0514 | 145 | 0.0659 | 0.8430 | 0.4544 | 0.2357 | 0.9106 |
| {Glocoseplasma2=high} => {InhospitalMortality=inhospital death} | 0.0500 | 141 | 0.0762 | 0.8197 | 0.4480 | 0.2500 | 0.8920 |
| {WBC2=high} => {InhospitalMortality=inhospital death} | 0.0461 | 130 | 0.0802 | 0.7558 | 0.4180 | 0.2462 | 0.8712 |
| {HigestCKMB2=high} => {InhospitalMortality=inhospital death} | 0.0426 | 120 | 0.0506 | 0.6976 | 0.3741 | 0.1880 | 0.9073 |
| {VENTOption=yes} => {InhospitalMortality=inhospital death} | 0.0202 | 57 | 0.6867 | 0.6867 | 0.5090 | 0.4770 | 0.4494 |
| {THTN=yes} => {InhospitalMortality=inhospital death} | 0.0340 | 96 | 0.0793 | 0.5581 | 0.3187 | 0.2104 | 0.8071 |
| {Education=illiterate} => {InhospitalMortality=inhospital death} | 0.0280 | 79 | 0.0925 | 0.4593 | 0.2759 | 0.2061 | 0.7201 |
| {mainfourcat_sm=four} => {InhospitalMortality=inhospital death} | 0.0284 | 80 | 0.1834 | 0.4651 | 0.3243 | 0.2921 | 0.5 |

### 5-2 Examining the effectiveness of some frequently used drugs in preventing or reducing In-Hospital Mortality
In this section, the effects of some prominent and widely used drugs on in-hospital mortality are investigated. This study aimed to determine how the patients' survival was affected by their medications both before and during hospitalization. Looking at table 7 (this table is sorted based on the frequency of repetition, count), it can be seen that among the medication during hospitalization, Aspirin, Statin, Clopidogrel, ACEinhibitors, and PPIs are the most common medications that have a very high effect on saving Patients have died (by observing the max confidence value). Moreover, the rules related to these medications have extremely high levels of confidence (max confidence values of 99.77%, 99.32%, 99.43%, 95.72%, and 93.99%, respectively). Additionally, among the medications taken prior to hospitalization, we can list Aspirin, ARBs, Beta Blockers, Statins, and ACEinhibitors, which are the most popular and most effective medications among patients (with max confidence values of 94.40% and 94.72%, 94.17%, 94.68%, and 92.52%, respectively).

**Table 7: The most important medications used by patients individually during hospitalization and prior to hospitalization, and their relationship with In-Hospital Mortality**

| Rules (X → Y) | Support | count | confidence | Max Confidence | Kulczynski | cosine | imbalance |
|---|---|---|---|---|---|---|---|
| {AspirinHos=yes} => {InhospitalMortality=alive} | 0.9368 | 2638 | 0.9398 | 0.9977 | 0.9688 | 0.9683 | 0.0579 |
| {StatinsHos=yes} => {InhospitalMortality=alive} | 0.9325 | 2626 | 0.9405 | 0.9932 | 0.9669 | 0.9665 | 0.0527 |
| {ClopidogrelHos=yes} => {InhospitalMortality=alive} | 0.9148 | 2576 | 0.9401 | 0.9743 | 0.9572 | 0.9571 | 0.0342 |
| {ACEinhibitorsHos=yes} => {InhospitalMortality=alive} | 0.8736 | 2460 | 0.9572 | 0.9572 | 0.9438 | 0.9437 | 0.0269 |
| {PPIsHos=yes} => {InhospitalMortality=alive} | 0.8604 | 2423 | 0.9399 | 0.9399 | 0.9281 | 0.9281 | 0.0236 |
| {BetaBlockersHos=yes} => {InhospitalMortality=alive} | 0.8572 | 2414 | 0.9549 | 0.9549 | 0.9340 | 0.9337 | 0.0421 |
| {heparinunfracHos=yes} => {InhospitalMortality=alive} | 0.6893 | 1941 | 0.9381 | 0.9381 | 0.8361 | 0.8299 | 0.2074 |
| {heparinLWHos=yes} => {InhospitalMortality=alive} | 0.4158 | 1171 | 0.9590 | 0.9590 | 0.7010 | 0.6517 | 0.5282 |
| {EptifibatideHos=yes} => {InhospitalMortality=alive} | 0.3825 | 1077 | 0.9565 | 0.9565 | 0.6819 | 0.6242 | 0.5637 |
| {DiureticsHos=yes} => {InhospitalMortality=alive} | 0.2443 | 688 | 0.9113 | 0.9113 | 0.5857 | 0.4869 | 0.6968 |
| {Aspirin=yes} => {InhospitalMortality=alive} | 0.2216 | 624 | 0.9440 | 0.9440 | 0.5900 | 0.4720 | 0.7396 |
| {MRAsHos=yes} => {InhospitalMortality=alive} | 0.2021 | 569 | 0.9563 | 0.9563 | 0.5858 | 0.4537 | 0.7674 |
| {ARBs=yes} => {InhospitalMortality=alive} | 0.1974 | 556 | 0.9472 | 0.9472 | 0.5787 | 0.4463 | 0.7690 |
| {BetaBlockers=yes} => {InhospitalMortality=alive} | 0.1548 | 436 | 0.9417 | 0.9417 | 0.5533 | 0.3941 | 0.8165 |
| {Statins=yes} => {InhospitalMortality=alive} | 0.1200 | 338 | 0.9468 | 0.9468 | 0.5373 | 0.3479 | 0.8588 |
| {ACEinhibitors=yes} => {InhospitalMortality=alive} | 0.0835 | 235 | 0.9252 | 0.9252 | 0.5070 | 0.2868 | 0.8975 |
| {ARBsHos=yes} => {InhospitalMortality=alive} | 0.0763 | 215 | 0.9471 | 0.9471 | 0.5142 | 0.2775 | 0.9100 |

Aspirin is the most crucial medicine in this trial for preventing death, as seen in Table 7. Table 8 provides a thorough assessment of this medication. It is clear that using this medication both before and during hospitalization can significantly reduce the number of fatalities; among the 661 patients who took aspirin prior to hospitalization, only 37 people passed away, and this rate is 169 for 2807 patients while in the hospital. In addition, there has been a considerable reduction in the risk of in-hospital death for patients who have used this medication both before and during hospitalization. In other words, just 36 of the 658 patients who used this medication prior to and during hospitalization died.

**Table 8: The impact of aspirin before and after hospitalization on In-Hospital Mortality**

| Rules (X → Y) | Support | count | confidence | Max Confidence | Kuczynski | cosine | imbalance |
|---|---|---|---|---|---|---|---|
| {AspirinHos=yes} => {InhospitalMortality=alive} | 0.9368 | 2638 | 0.9398 | 0.9977 | 0.9688 | 0.9683 | 0.0579 |
| {AspirinHos=yes} => {InhospitalMortality=inhospital death} | 0.0600 | 169 | 0.0602 | 0.9826 | 0.5214 | 0.2432 | 0.9377 |
| {Aspirin=yes,AspirinHos=yes} => {InhospitalMortality=alive} | 0.2209 | 622 | 0.9453 | 0.9453 | 0.5903 | 0.4716 | 0.7410 |
| {Aspirin=yes} => {InhospitalMortality=alive} | 0.2216 | 624 | 0.9440 | 0.9440 | 0.5900 | 0.4720 | 0.7396 |
| {Aspirin=yes} => {InhospitalMortality=inhospital death} | 0.0131 | 37 | 0.0560 | 0.2151 | 0.1355 | 0.1097 | 0.6143 |
| {Aspirin=yes,AspirinHos=yes} => {InhospitalMortality=inhospital death} | 0.0128 | 36 | 0.0547 | 0.2093 | 0.1320 | 0.1070 | 0.6121 |

As shown in Table 9, the impact of multiple medications taken to a patient simultaneously both before and during hospitalization has also been studied in order to learn more about the medications utilized. For instance, 2621 hospitalized patients who took the aspirin and statin did not experience any bad outcomes (maximum confidence = 99.13%). For the aforementioned rule, the value of Kulczynski is extremely high (0.96), indicating the two-sided relationship of this law. Alternatively, all 70 patients who got aspirin, eptifibatideHos, beta-blockersHos, and ARBs lived. The effectiveness of the medications in lowering the risk of bad outcomes can be seen by looking at the other rules of this table.

**Table 9: The most important medications used by patients simultaneously during hospitalization and before hospitalization, and their relationship with In-Hospital Mortality**

| Rules (X → Y) | Support | count | confidence | Max Confidence | Kulczynski | cosine | imbalance |
|---|---|---|---|---|---|---|---|
| {Aspirin=yes,EptifibatideHos=yes,ARBs=yes} => {InhospitalMortality=alive} | 0.0284 | 80 | 1 | 1 | 0.5151 | 0.1739 | 0.9697 |
| {heparinunfracHos=yes,ACEinhibitorsHos=yes,ARBs=yes,Diuretics=yes} => {InhospitalMortality=alive} | 0.0167 | 47 | 1 | 1 | 0.5089 | 0.1333 | 0.9822 |
| {heparinunfracHos=yes,ARBsHos=yes,MRAsHos=yes,PPIsHos=yes} => {InhospitalMortality=alive} | 0.0135 | 38 | 1 | 1 | 0.5072 | 0.1199 | 0.9856 |
| {ACEinhibitorsHos=yes,ARBsHos=yes,MRAsHos=yes,PPIsHos=yes} => {InhospitalMortality=alive} | 0.0156 | 44 | 1 | 1 | 0.5083 | 0.1290 | 0.9834 |

| Rules (X → Y) | Support | count | confidence | Max Confidence | Kulczynski | cosine | imbalance |
|---|---|---|---|---|---|---|---|
| {Aspirin=yes,ACEinhibitors=yes,ACEinhibitorsHos=yes,DiureticsHos=yes} => {InhospitalMortality=alive} | 0.0128 | 36 | 1 | 1 | 0.5068 | 0.1167 | 0.9864 |
| {Aspirin=yes,heparinLWHos=yes,ACEinhibitors=yes,ACEinhibitorsHos=yes} => {InhospitalMortality=alive} | 0.0149 | 42 | 1 | 1 | 0.5079 | 0.1260 | 0.9841 |
| {heparinunfracHos=yes,BetaBlockers=yes,ARBs=yes,Statins=yes} => {InhospitalMortality=alive} | 0.0131 | 37 | 1 | 1 | 0.5070 | 0.1183 | 0.9860 |
| {EptifibatideHos=yes,ACEinhibitorsHos=yes,ARBs=yes,Statins=yes} => {InhospitalMortality=alive} | 0.0149 | 42 | 1 | 1 | 0.5079 | 0.1260 | 0.9841 |
| {heparinunfracHos=yes,ACEinhibitorsHos=yes,ARBs=yes,Statins=yes} => {InhospitalMortality=alive} | 0.0302 | 85 | 1 | 1 | 0.5161 | 0.1793 | 0.9679 |
| {heparinunfracHos=yes,heparinLWHos=yes,BetaBlockersHos=yes,Statins=yes} => {InhospitalMortality=alive} | 0.0241 | 68 | 1 | 1 | 0.5129 | 0.1604 | 0.9743 |
| {Aspirin=yes,EptifibatideHos=yes,heparinunfracHos=yes,ARBs=yes} => {InhospitalMortality=alive} | 0.0281 | 79 | 1 | 1 | 0.5149 | 0.1729 | 0.9701 |
| {Aspirin=yes,EptifibatideHos=yes,BetaBlockersHos=yes,ARBs=yes} => {InhospitalMortality=alive} | 0.0249 | 70 | 1 | 1 | 0.5132 | 0.1627 | 0.9735 |
| {Aspirin=yes,EptifibatideHos=yes,ARBs=yes,PPIsHos=yes} => {InhospitalMortality=alive} | 0.0263 | 74 | 1 | 1 | 0.5140 | 0.1673 | 0.9720 |
| {Aspirin=yes,EptifibatideHos=yes,ACEinhibitorsHos=yes,ARBs=yes} => {InhospitalMortality=alive} | 0.0256 | 72 | 1 | 1 | 0.5136 | 0.1650 | 0.9728 |
| {Aspirin=yes,ClopidogrelHos=yes,EptifibatideHos=yes,ARBs=yes} => {InhospitalMortality=alive} | 0.0281 | 79 | 1 | 1 | 0.5149 | 0.1729 | 0.9701 |
| {Aspirin=yes,EptifibatideHos=yes,ARBs=yes,StatinsHos=yes} => {InhospitalMortality=alive} | 0.0284 | 80 | 1 | 1 | 0.5151 | 0.1739 | 0.9697 |
| {Aspirin=yes,AspirinHos=yes,EptifibatideHos=yes,ARBs=yes} => {InhospitalMortality=alive} | 0.0284 | 80 | 1 | 1 | 0.5151 | 0.1739 | 0.9697 |
| {AspirinHos=yes,StatinsHos=yes} => {InhospitalMortality=alive} | 0.9308 | 2621 | 0.9404 | 0.9913 | 0.9659 | 0.9655 | 0.0509 |
| {Aspirin=yes,EptifibatideHos=yes,BetaBlockersHos=yes,PPIsHos=yes} => {InhospitalMortality=alive} | 0.0689 | 194 | 0.9898 | 0.9898 | 0.5316 | 0.2695 | 0.9252 |
| {Aspirin=yes,heparinLWHos=yes,ACEinhibitorsHos=yes,ARBs=yes} => {InhospitalMortality=alive} | 0.0316 | 89 | 0.9889 | 0.9889 | 0.5113 | 0.1824 | 0.9656 |
| {heparinunfracHos=yes,BetaBlockers=yes,ACEinhibitorsHos=yes,Statins=yes} => {InhospitalMortality=alive} | 0.0298 | 84 | 0.9882 | 0.9882 | 0.5100 | 0.1772 | 0.9675 |
| {Aspirin=yes,ACEinhibitorsHos=yes,ARBs=yes,Statins=yes} => {InhospitalMortality=alive} | 0.0288 | 81 | 0.9878 | 0.9878 | 0.5092 | 0.1740 | 0.9686 |
| {Aspirin=yes,EptifibatideHos=yes,BetaBlockersHos=yes,Statins=yes} => {InhospitalMortality=alive} | 0.0270 | 76 | 0.9870 | 0.9870 | 0.5079 | 0.1684 | 0.9705 |
| {Aspirin=yes,EptifibatideHos=yes,BetaBlockersHos=yes} => {InhospitalMortality=alive} | 0.0771 | 217 | 0.9864 | 0.9864 | 0.5342 | 0.2845 | 0.9158 |
| {Aspirin=yes,AspirinHos=yes,EptifibatideHos=yes,BetaBlockersHos=yes} => {InhospitalMortality=alive} | 0.0767 | 216 | 0.9863 | 0.9863 | 0.5340 | 0.2839 | 0.9161 |
| {Aspirin=yes,ClopidogrelHos=yes,EptifibatideHos=yes,BetaBlockersHos=yes} => {InhospitalMortality=alive} | 0.0760 | 214 | 0.9862 | 0.9862 | 0.5336 | 0.2825 | 0.9169 |
| {Aspirin=yes,EptifibatideHos=yes,BetaBlockersHos=yes,StatinsHos=yes} => {InhospitalMortality=alive} | 0.0760 | 214 | 0.9862 | 0.9862 | 0.5336 | 0.2825 | 0.9169 |
| {Aspirin=yes,EptifibatideHos=yes,heparinunfracHos=yes,BetaBlockersHos=yes} => {InhospitalMortality=alive} | 0.0753 | 212 | 0.9860 | 0.9860 | 0.5331 | 0.2812 | 0.9176 |

## 6- Conclusion

In this paper, a novel framework called GWO-FI was introduced to diagnose in-hospital mortality. It combines the gray wolf algorithm with itemsets. Frequent items were initially extracted by Apriori and applied as features to the original dataset before using the gray wolf technique to select the best subset of features. The diagnostic model's results showed satisfactory performance with an auroc value of 0.9734. In addition, the second phase of the Apriori algorithm was used to analyze the most important factors affecting mortality in the hospital as well as the duration of hospitalization. According to the results of this section, the use of aspirin before and during hospitalization was the best medicine. Low ejection fraction, high CPK, improper heart efficiency (WorstKLLIPclass), and old age were also listed as the primary causes of hospital deaths.

## Declarations

## Ethics approval and consent to participate

Our project was approved by the Kermanshah University of Medical Sciences (IR.KUMS.REC.1400.252)

**Availability of data and materials**
The data of this study are available from the corresponding author upon reasonable request.

**Conflict of interest:**
The authors declare that they have no conflicts of interest.
**Acknowledgment**
This study was funded by grant number 95449 from Kermanshah University of Medical Sciences and Health Services, Kermanshah. Iran.

**Reference**

[1] World Heart Organization (WHO). Cardiovascular disease. https://www.who.int/cardiovascular_diseases/about_cvd/en/ Accessed on 25 Dec 2022.

[2] A. V. Mattioli, M. Ballerini Puviani, M. Nasi, and A. Farinetti, "COVID-19 pandemic: the effects of quarantine on cardiovascular risk," (in eng), *Eur J Clin Nutr,* vol. 74, no. 6, pp. 852-855, Jun 2020.

[3] J. Han, M. Kamber, and J. Pei, *Data Mining: Concepts and Techniques*. Morgan Kaufmann Publishers Inc., 2011, p. 696.

[4] H. Cheng, X. Yan, J. Han, and P. S. Yu, "Direct Discriminative Pattern Mining for Effective Classification," in *2008 IEEE 24th International Conference on Data Engineering*, 2008, pp. 169-178.

[5] L. I. Veldhuis, N. J. C. Woittiez, P. W. B. Nanayakkara, and J. Ludikhuize, "Artificial Intelligence for the Prediction of In-Hospital Clinical Deterioration: A Systematic Review," vol. 4, no. 9, p. e0744, 2022.

[6] S. Iwase *et al.*, "Prediction algorithm for ICU mortality and length of stay using machine learning," *Scientific Reports,* vol. 12, no. 1, p. 1.2022 28/07/2022, 2912

[7] Y.-w. Chen *et al.*, "Learning to predict in-hospital mortality risk in the intensive care unit with attention-based temporal convolution network," *BMC Anesthesiology,* vol. 22, no. 1, p. 119, 2022/04/23 2022.

[8] D. A. Alabbad *et al.*, "Machine learning model for predicting the length of stay in the intensive care unit for Covid-19 patients in the eastern province of Saudi Arabia," (in eng), *Inform Med Unlocked,* vol. 30, p. 100937, 2022.

[9] A. Yavari, A. Rajabzadeh, and F. J. a. e.-p. Abdali-Mohammadi, "A Profile-Based Binary Feature Extraction Method Using Frequent Itemsets for Improving Coronary Artery Disease Diagnosis," p. arXiv:2109.10966Accessed on: September 01, 2021Available: https://ui.adsabs.harvard.edu/abs/2021arXiv210910966Y

[10] A. Yavari, A. Rajabzadeh, and F. Abdali-Mohammadi, "Profile-based assessment of diseases affective factors using fuzzy association rule mining approach: A case study in heart diseases," *Journal of Biomedical Informatics,* vol. 116, p. 103695, 2.2021 /01/04/021

[11] S. Soffer, E. Klang, Y. Barash, E. Grossman, and E. Zimlichman, "Predicting In-Hospital Mortality at Admission to the Medical Ward: A Big-Data Machine Learning Model," *The American Journal of Medicine,* vol. 134, no. 2, pp. 227-234.e4.2021 /01/02/2021 ,

[12] A. Naemi, T. Schmidt, M. Mansourvar, M. Naghavi-Behzad, A. Ebrahimi, and U. K. Wiil, "Machine learning techniques for mortality prediction in emergency departments: a systematic review," *BMJ Open,* vol. 11, no. 11, p. e052663, 2021.

[13] S. Mostafa, O. Azam, and K.-A. Hadi, "Comparing of Data Mining Techniques for Predicting in-Hospital Mortality Among Patients with COVID-19," *Journal of Biostatistics and Epidemiology,* vol. 7, no. 2, 07/04 2021.

[14] B. Mahboub, M. T. A. Bataineh, H. Alshraideh, R. Hamoudi, L. Salameh, and A. Shamayleh, "Prediction of COVID-19 Hospital Length of Stay and Risk of Death Using Artificial Intelligence-Based Modeling," (in eng), *Front Med (Lausanne),* vol. 8, p. 592336, 2021.

[15] W. S. Kang *et al.*, "Artificial intelligence to predict in-hospital mortality using novel anatomical injury score," *Scientific Reports,* vol. 11, no. 1, p. 23534, 2021/12/07 2021.

[16] A. Gupta, R. Kumar, H. S. Arora, and B. Raman, "C-CADZ: computational intelligence system for coronary artery disease detection using Z-Alizadeh Sani dataset," *Applied Intelligence,* 2021/06/12 2021.

[17] J. Chrusciel, F. Girardon, L. Roquette, D. Laplanche, A. Duclos, and S. Sanchez, "The prediction of hospital length of stay using unstructured data," *BMC Medical Informatics and Decision Making,* vol. 21, no. 1, p. 351, 2021/12/18 2021.

[18] B. Aken, J.-M. Papaioannou, M. Mayrdorfer, K. Budde, F. A. Gers, and A. Löser, *Clinical Outcome Prediction from Admission Notes using Self-Supervised Knowledge Integration*. arXiv, 2021.

[19] R. N. Mekhaldi, P. Caulier, S. Chaabane, A. Chraibi, and S. Piechowiak, "Using Machine Learning Models to Predict the Length of Stay in a Hospital Setting," in *Trends and Innovations in Information Systems and Technologies*, Cham, 2020, pp. 202-211: Springer International Publishing.

[20] M. M. Ghiasi, S. Zendehboudi, and A. A. Mohsenipour, "Decision tree-based diagnosis of coronary artery disease: CART model," *Computer Methods and Programs in Biomedicine,* vol. 192, p. 105400, 2020/08/01/2020.

[21] Y.-G. Seo et al., "Cardiovascular disease risk factor clustering in children and adolescents: a prospective cohort study," *Archives of Disease in Childhood,* vol. 103, no. 10, p. 968, 2018.

[22] R. Alizadehsani et al., "Non-invasive detection of coronary artery disease in high-risk patients based on the stenosis prediction of separate coronary arteries," *Computer Methods and Programs in Biomedicine,* vol. 162, pp. 119-127, 2018.

[23] M. Tayefi et al., "hs-CRP is strongly associated with coronary heart disease (CHD): A data mining approach using decision tree algorithm," *Computer Methods and Programs in Biomedicine,* vol. 141, pp. 105-109, 2017.

[24] J. Domienik-Karłowicz et al., "Fourth universal definition of myocardial infarction. Selected messages from the European Society of Cardiology document and lessons learned from the new guidelines on ST-segment elevation myocardial infarction and non-ST-segment elevation-acute coronary syndrome," (in eng), *Cardiol J,* vol. 28, no. 2, pp. 195-201, 2021.

[25] R. Alizadehsani et al., "A data mining approach for diagnosis of coronary artery disease," *Computer Methods and Programs in Biomedicine,* vol. 111, no. 1, pp. 52-61, 2013.

[26] A. S. Levey et al., "A new equation to estimate glomerular filtration rate," (in eng), *Ann Intern Med,* vol. 150, no. 9, pp. 604-12, May 5 2009.

[27] S. Mirjalili, S. M. Mirjalili, and A. Lewis, "Grey Wolf Optimizer," *Advances in Engineering Software,* vol. 69, pp. 46-61, 2014/03/01/ 2014.

[28] E. Emary, H. M. Zawbaa, and A. E. Hassanien, "Binary grey wolf optimization approaches for feature selection," *Neurocomputing,* vol. 172, pp. 371-381, 2016/01/08/ 2016.

[29] R. Agrawal and R. Srikant, "Fast Algorithms for Mining Association Rules " in *Proceedings of the 20th International Conference on Very Large Data Bases*, 1994, pp. 487-499.